\documentclass[conference]{IEEEtran}
\IEEEoverridecommandlockouts
\usepackage{cite}
\usepackage{float}
\usepackage{amsmath,graphicx,amsfonts,amssymb,url}
\usepackage{setspace}
\usepackage[caption=false]{subfig}
\usepackage{blindtext}
\usepackage[inline]{enumitem}
\usepackage{color}
 \usepackage{subfig}
\usepackage{multirow}
\newsavebox{\measurebox}
\usepackage{minibox}
\usepackage{array,multirow,booktabs}
\usepackage{graphicx}
\usepackage{siunitx}
\usepackage{amsfonts,amssymb}
\usepackage{balance}
\usepackage{pifont}% http://ctan.org/pkg/pifont

\usepackage{algorithm}
\usepackage{algorithmic}
\usepackage{authblk}
\newcommand*{\affaddr}[1]{#1} 
\newcommand*{\affmark}[1][*]{\textsuperscript{#1}}
\newcommand*{\email}[1]{\texttt{#1}}

\newcommand{\suiyi}[1]{\textcolor{black}{#1}}
\newcommand{\suiyinew}[1]{\textcolor{black}{#1}}
\newcommand{\modification}[1]{\textcolor{black}{#1}}

\def\BibTeX{{\rm B\kern-.05em{\sc i\kern-.025em b}\kern-.08em
    T\kern-.1667em\lower.7ex\hbox{E}\kern-.125emX}}
\begin{document}

\title{\suiyi{Few-Shot Object Detection in Real Life: \\ Case Study on Auto-Harvest}\\
}

\author{Kevin Riou\affmark[1], Jingwen Zhu\affmark[1], Suiyi Ling\affmark[1]\affmark[2], Mathis Piquet\affmark[1],Vincent Truffault\affmark[3], Patrick Le Callet\affmark[1]\\
\affaddr{\affmark[1]University of Nantes \ }
\affaddr{\affmark[2]Capacites SAS \ }
\affaddr{\affmark[3]CTIFL \ }
\\
\email{\tt\small  \{kevin.riou, jingwen.zhu, mathis.piquet\}@etu.univ-nantes.fr;}
\\
\email{\tt\small \{suiyi.ling, patrick.lecallet\}@univ-nantes.fr} 
\email{\tt\small Vincent@futuragaia.com}
}

\IEEEoverridecommandlockouts
\IEEEpubid{\makebox[\columnwidth]{978-1-7281-9320-5/20/\$31.00~\copyright2020 European Union \hfill} \hspace{\columnsep}\makebox[\columnwidth]{ }}

\maketitle
 \IEEEpubidadjcol
\begin{abstract}
\suiyi{Confinement during COVID-19 has caused serious effects on agriculture all over the world. As one of the efficient solutions, mechanical harvest/auto-harvest that is based on object detection and robotic harvester becomes an urgent need. Within the auto-harvest system, robust few-shot object detection model is one of the bottlenecks, since the system is required to deal with new vegetable/fruit categories and the collection of large-scale annotated datasets for all the novel categories is expensive. There are many few-shot object detection models that were developed by the community. Yet whether they could be employed directly for real life agricultural applications is still questionable, as there is a context-gap between the commonly used training datasets and the images collected in real life agricultural scenarios. To this end, in this study, we present a novel cucumber dataset and propose two data augmentation strategies that help to bridge the context-gap.  Experimental results show that 1) the state-of-the-art few-shot object detection model performs poorly on the novel `cucumber' category;  and 2) the proposed augmentation strategies outperform the commonly used ones.  }
\end{abstract}


\section{Introduction}
\label{sec:intro}
 
% 1. background of harvesting, why we need few-shot for it 
\suiyi{COVID-19 has brought an extremely painful period for the world. Apart from continuing to take its toll, the pandemic has also spawned a great number of economic losses. Especially in agriculture, due to the confinement, harvesting slowly becomes one of the enormous challenges. Therefore, the development of robust `auto-harvesting' is of far greater urgency. The existing mechanical harvest/auto-harvest system is normally based on 1) auto-detection of the matured vegetables/fruits and 2) the auto-harvest using robots within a limited reachable zoom. In most of the auto-harvest systems, auto-detection of the target regions is usually the bottleneck and thus is more vital.  }

% 2. Few-shot models
\suiyi{Recent deep convolutionnal networks~\cite{huang2017speed,redmon2016you} have achieved significant improvement in object detection. Most of them depends heavily on large-scale training dataset like COCO~\cite{lin2014common} and PASCAL VOC~\cite{everingham2010pascal,everingham2015pascal}. Such dependency raises questions like 1) how to deal with the new categories that was not involved in the training set; and more importantly, 2) how to employ them in real life, when there are no enough samples for target categories. For auto-detection of vegetables/fruits, it is often required to deal with new categories with limited training samples per category as samples and annotations could be difficult and expensive to collect. }

\suiyi{To tackle such intractable problem, recently, few-shot~\cite{vinyals2016matching,snell2017prototypical} and meta learning~\cite{finn2017model} based models were proposed. Most of these models target to obtain a model that is able to address both base and novel categories at test time. In general, there are two training phases/stages within the training scheme of these models, including 1) a base learner training or meta training stage where the base categories are utilized; 2) a fine-tuner/task adaptor training or meta testing stage, where the new categories are used. This type of training-testing schemes target to transfer the task-relevant (recognition/detection) knowledge from the base categories to the novel ones, and more importantly force the model to learn with only a few samples. }

\begin{figure}[t]
\centering
 \mbox{ \parbox{1\textwidth}{
   \begin{minipage}[b]{0.16\textwidth}
   \subfloat[Normal exposure ]
  {\label{ }\includegraphics[width=\textwidth]{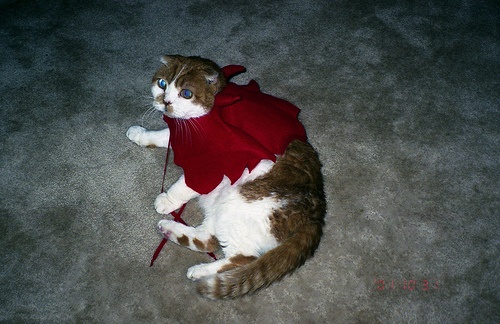}}  \end{minipage}
  \begin{minipage}[b]{0.16\textwidth}
  \subfloat[Clean background ]
  {\label{ }\includegraphics[width=\textwidth]{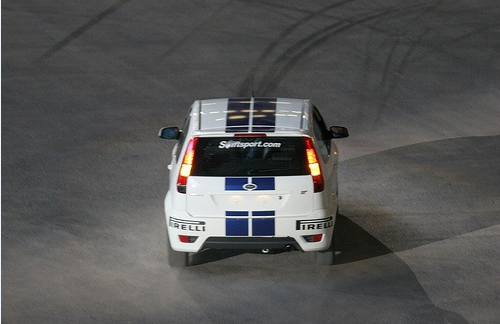}}
  \end{minipage}
  \begin{minipage}[b]{0.16\textwidth}
  \subfloat[No Occlusion]
  {\label{ }\includegraphics[width=\textwidth]{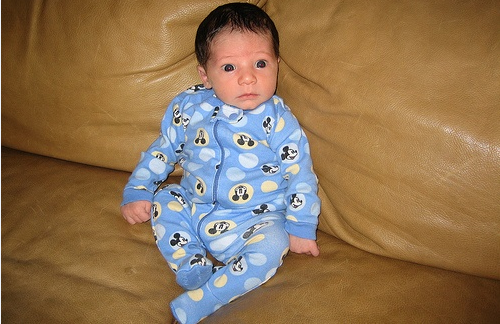}}
  \end{minipage}
  \\
   \begin{minipage}[b]{0.16\textwidth}
   \subfloat[Overexposure]
  {\label{ }\includegraphics[width=\textwidth]{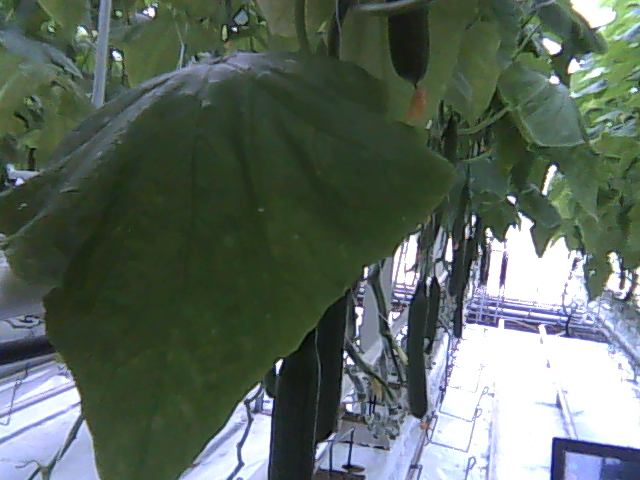}}  \end{minipage}
  \begin{minipage}[b]{0.16\textwidth}
  \subfloat[Complex background]
  {\label{ }\includegraphics[width=\textwidth]{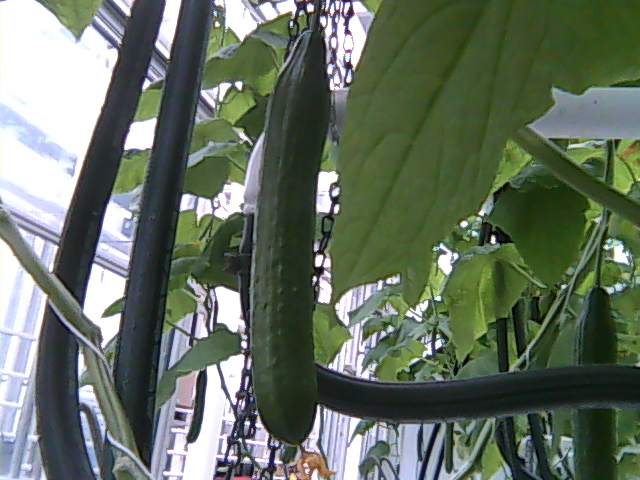}}
  \end{minipage}
  \begin{minipage}[b]{0.16\textwidth}
  \subfloat[Occlusion]
  {\label{ }\includegraphics[width=\textwidth]{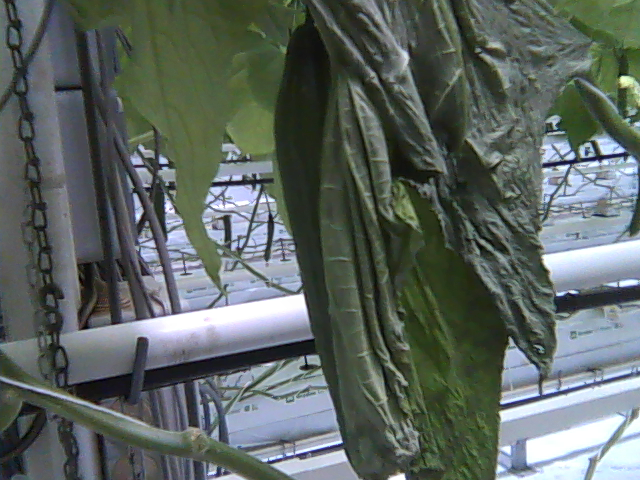}}
  \end{minipage}
   }}
\caption{\suiyi{Examples explaining the `Gap' between the training dataset used in the literature and target real life problem. \textbf{First row:} images in commonly used PASCAL dataset; \textbf{Second row:} Images from our cucumber dataset. }}
\label{teaser}
\end{figure}

% 3. Weakness of current few-shot detection methods and raise challenges in real life scenarios 
\suiyi{Nevertheless, real life few-shot problems are more challenging due to factors like noisy imaging conditions~\cite{ling2020few}. In the case of auto vegetables/fruits detection, we notice that there is an obvious gap between the image-context of dataset used to train the state-of-the-art few-shot object detection models and the ones of real life vegetable images collected in greenhouse. Examples are depicted in Fig.~\ref{teaser}. 
As shown, the aforementioned challenges include 1) \textbf{image conditions:} as images are not taken under controlled camera setting, it is common to see samples collected in real life with worse image conditions, \textsl{e.g.}, overexposure as shown in Fig.~\ref{teaser} (d);  2) \textbf{not all the objects appear in the image are the target:} since auto harvest relies on the robots to harvest,  only objects that are within a reachable area are interesting, \textsl{e.g.} as shown in Fig.~\ref{teaser} (e), only the cucumbers shown in the foreground are the targets; 3) \textbf{noisy background and occlusion:} images taken in certain real life scenarios could have a complex and noisy background, and the target objects could be occluded as shown in Fig.~\ref{teaser} (f). Thus, except for transferring the detection knowledge from the base categories to the new ones, the context-knowledge (\textsl{e.g.}, image conditions, background \textsl{etc.}) of new scenarios should also be transferred for real life applications. } 

% goal & contributions of the paper

\suiyi{In this study, based on the discussions above, we aim at re-visiting the state-of-the-art few-shot detection models for real life greenhouse applications, and exploring data augmentation strategies to better transfer context-knowledge of the new scenarios.  The main contributions of our paper are twofold: 1) A novel cucumber dataset with bounding boxes annotations is released; 2) Two data augmentation strategies are proposed to take the context of the new applications into account. }

\section{\suiyi{Related Works}}
\label{sec:RL}
% related work should be after  introduction
% Since our paper targets few-shot object , I suggest to combine zero-shot with few-shot

% To mention author, u just need to mention the family name of the first author
\suiyi{Few-shot object detection, including zero-shot object detection, aims to accurately detect novel category of objects that are not involved in the training procedure using few samples, \textsl{i.e,} shots, of the new category. Compared to the traditional recognition or object detection task, this brand new problem is significantly more challenging due to the ill-posed nature and inherent complexity of detecting absolutely unknown categories. To tackle this challenging problem, many works have been developed. Michaelis~\textsl{et al.}~\cite{michaelis2018one} improved the Mask R-CNN with a Siamese architecture. In~\cite{rahman2019transductive}, the author proposed a zero-shot model based on transductive learning. Recently, natural language descriptions were considered to better address the problem~\cite{li2019zero}. An R-CNN based re-weighting network was presented in  \cite{yan2019meta}, which disentangles multi-object information and turns Faster/Mask R-CNN into a meta-learner for few-shot object detection. Following a similar recipe, Kang \textsl{et al.}  proposed a region proposal free re-weighting network based on YOLOv2. } 

\begin{figure}[t]
\centering
\mbox{\parbox{1\textwidth}{
 \begin{minipage}[b]{0.24\textwidth}
 \subfloat[]
 {\label{ }\includegraphics[width=\textwidth]{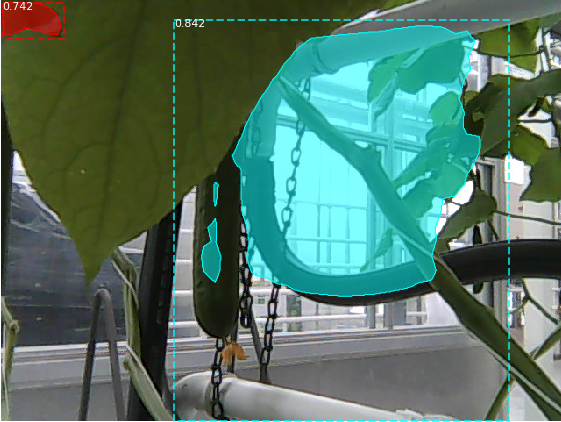}} \end{minipage}
 \begin{minipage}[b]{0.24\textwidth}
 \subfloat[]
 {\label{ }\includegraphics[width=\textwidth]{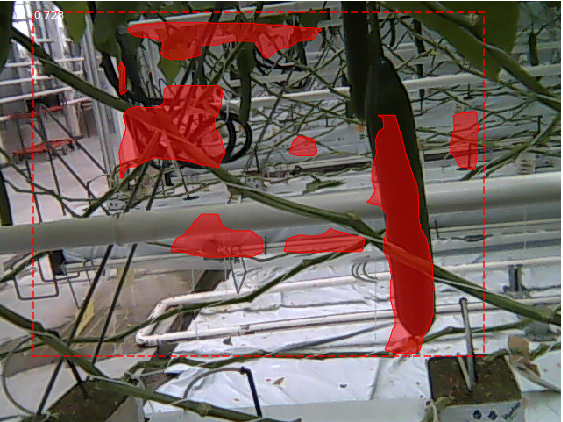}} \end{minipage}
}}
\caption{Result of Siamese Mask-R-CNN model on zero-shot cucumber segmentation.}
\label{resMaskRcnn}
\end{figure}

\suiyi{Nevertheless, most of the aforementioned models are trained and tested on the same type of datasets, \textsl{e.g.} the PASCAL dataset. The distributions of theses datasets could be significantly different compared to the images collected in real life scenarios. More specifically, there will be a large domain-shift between the source (datasets used to train existing models) and the target (real life applications) domains. Therefore, they are prone to fail when applied in typical real life scenarios. An example is depicted in Fig.~\ref{resMaskRcnn}, where the model proposed in ~\cite{michaelis2018one} incorrectly segments the background as the target `cucumber' regions.}

\section{\suiyi{Our Cucumber Dataset}}
To remedy the lack of real life dataset for automatic harvesting, we collected a new dataset that contains images of cucumbers in the greenhouse. The dataset is publicly available at \textsl{https://github.com/KevinRiou22/Labeled-cucumber-dataset}. Details of the dataset is summarized in Table~\ref{tab:data}.

 \suiyi{\textbf{Data collection:} The collection of the dataset was organized by \textsl{Centre Technique Inter-professionnel des Fruits et Légumes} (CTIFL)~\cite{CTIFLWebSite} and the images were taken within their greenhouse ($47^\circ 17'11.6''N  \hspace{0.2cm} 1^\circ 27'34.3''W$). CTIFL is an `inter-professional center for the fruits and vegetables'. Especially, it is a french agency that aims at developing the knowledge and expertise of all professions around fruits and vegetables sectors.}
 \begin{table}[htbp]
\caption{Summary of our cucumber dataset. }
\label{tab:data}
\centering
\begin{tabular}{|c|c|}  \hline
 Number of images &  555 \\ \hline
 Resolution & 640 x 480  \\ \hline
 Number of annotation per image & 1 - 4\\ \hline
\end{tabular}
\label{result}
\end{table}

\begin{figure}[t]
\centering
\mbox{\parbox{1\textwidth}{
 \begin{minipage}[b]{0.16\textwidth}
   \subfloat[]
  {\label{datasetA }\includegraphics[width=\textwidth]{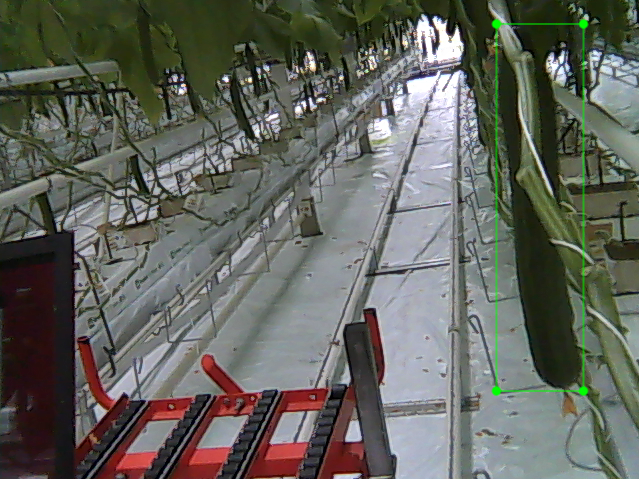}}  \end{minipage}
   \begin{minipage}[b]{0.16\textwidth}
    \subfloat[]
  {\label{ datasetB }\includegraphics[width=\textwidth]{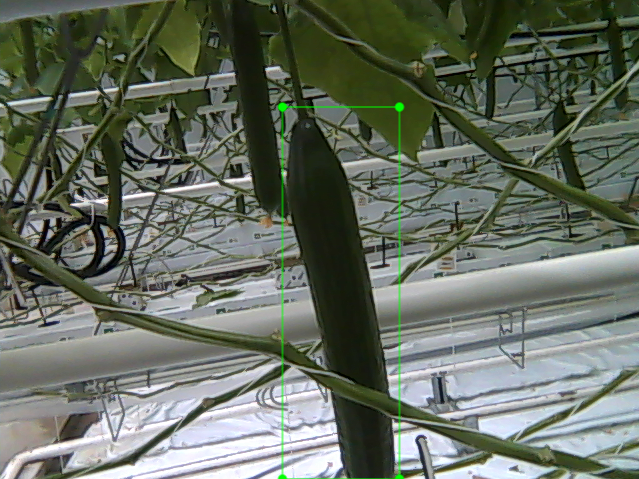}} \end{minipage}
   \begin{minipage}[b]{0.16\textwidth}
   \subfloat[]
   {\label{datasetC}\includegraphics[width=\textwidth]{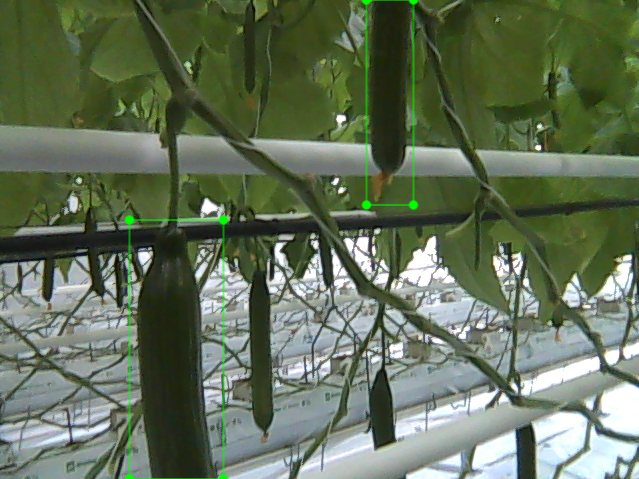}} \end{minipage}
   \\
   \begin{minipage}[b]{0.16\textwidth}
   \subfloat[]
   {\label{}\includegraphics[width=\textwidth]{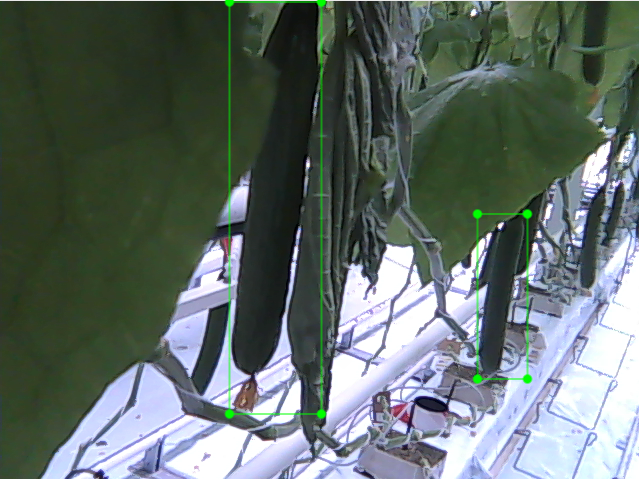}} \end{minipage}
   \begin{minipage}[b]{0.16\textwidth}
   \subfloat[]
   {\label{}\includegraphics[width=\textwidth]{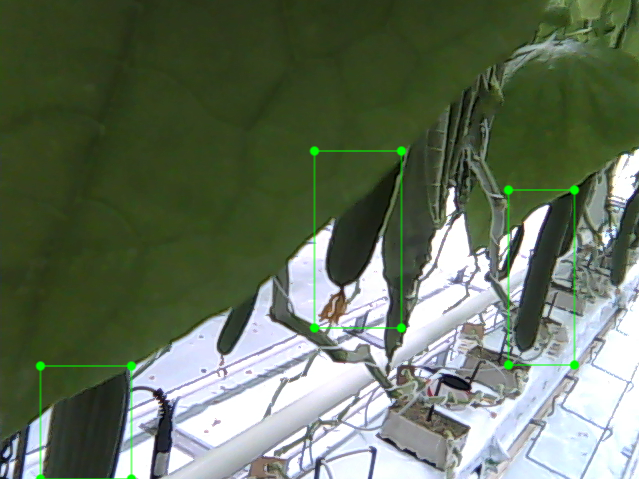}} \end{minipage}
   \begin{minipage}[b]{0.16\textwidth}
   \subfloat[]
   {\label{}\includegraphics[width=\textwidth]{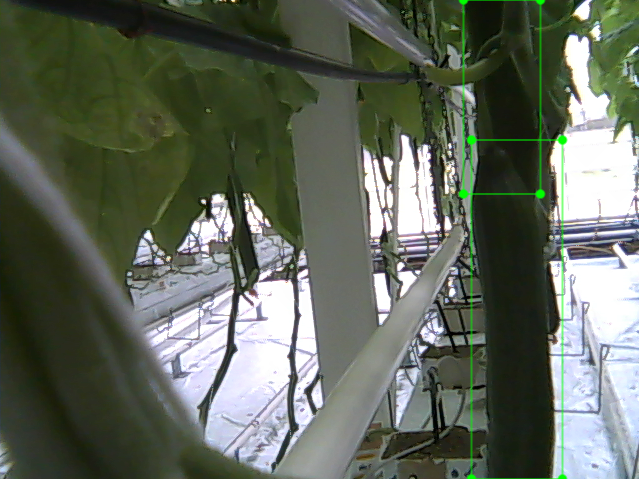}} \end{minipage}
   }}
\caption{\suiyi{Examples from our dataset, where the green boxes are the labels indicating the location of the cucumbers: (a) shows the trolley that carried the camera along the rails in the greenhouse; (b,c) are examples taken facing directly against the `cucumber rows' as depicted in (a); (d, c, e) are examples collected with different camera angles.}}
\label{dataset}
\end{figure}

 \suiyi{In their greenhouses, the plants, \textsl{i.e.,} the cucumbers,  were planted in rows. Between every two rows, rails were set up to allow trolleys or automatic harvesting robots to forward and harvest the ripe vegetables/fruits automatically. To facilitate the automation of harvesting, cameras were mounted on the trolley, around 80cm away from the plants as shown in Fig.~\ref{dataset} (a). The angles of the cameras are adjustable. To avoid `motion blur', the trolleys stopped constantly along their moving trajectory to take images from different angles. In general, images of plants were taken in front of the rows,~\textsl{e.g.}, Fig.~\ref{dataset} (b-d). Additionally, we also enriched the dataset by varying the cameras' angles, as presented in Fig.~\ref{dataset} (e-f). It is worth emphasizing that only the row closest to the trolley is interesting, as the automatic robot targets only to harvest the plants (cucumbers) within a constraint surrounding range.}

\suiyi{ \textbf{Annotations}: When annotating the locations of target cucumbers, we followed the instructions provided by ~\cite{everingham2010pascal}, and thus followed a `PASCAL-VOC' format.  Only the cucumbers from the closest row to the trolley, \textsl{i.e.,} those in the foreground, were labeled, as they are within the accessible range of the robotic-arm of the trolley.}

\section{\suiyi{Re-visit state-of-the-art few-shot object detection models for real life scenario}}
\label{sec:model}

\suiyi{As summed up in Section \ref{sec:RL}, the few-shot object detection model proposed in~\cite{kang2019few}, namely FS-FRW, is one of the state-of-the-art models and achieves the best performances. Therefore, in this study, we re-visit and adapt this approach to few-shot cucumber detection in real life scenarios.}

\begin{figure*}[t]
\begin{center}
\includegraphics[width=18cm]{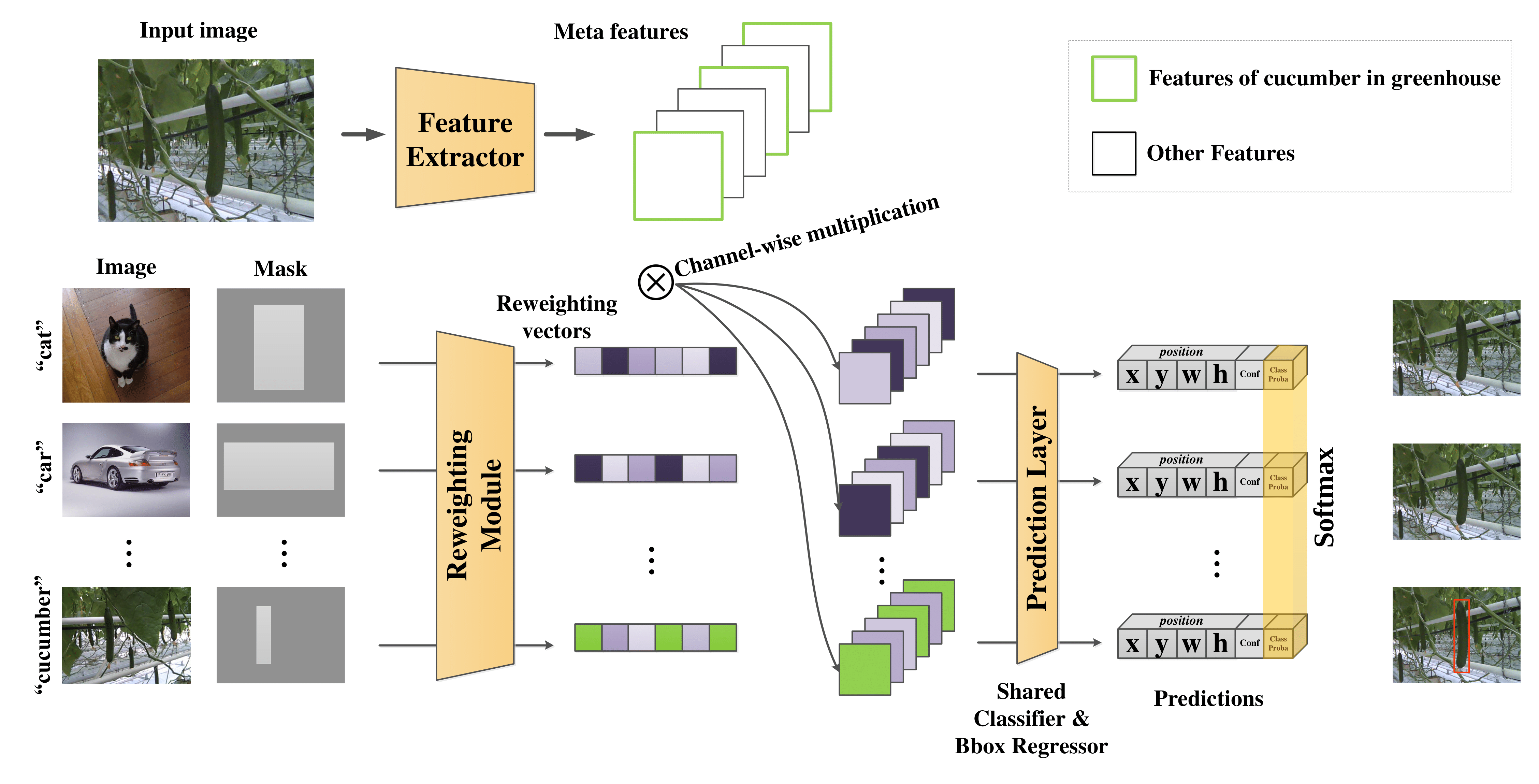}
\end{center}
   \caption{\suiyi{The overall framework of the adapted few-show object detection model based on feature re-weighting in greenhouse scenario~\cite{kang2019few}.}}
\label{structureReweighting}
\end{figure*}

\suiyi{The overall adapted FS-FRW is summarized in Fig.~\ref{structureReweighting}. The framework is composed of two main parts, including (1) a generalized feature learner $D$ that is trained to detect objects of novel categories using large-scale training samples per class, as shown in the upper part of Fig.~\ref{structureReweighting}; and (2) a feature re-weighter $M$ that reveals the vital meta-features, \textsl{i.e.,} vital components of the feature learner, and benefits the accurate identification/detection of objects belonging to a new category with only a few samples of this new category, as shown in the lower part of Fig.~\ref{structureReweighting}. They are trained together in an end-to-end and few-shot fashion so that novel categories of objects could be well detected.}

\suiyi{In this study, `cucumber' is considered as the novel category, and the goal is to accurately detect them utilizing FS-FRW under a complex, noisy greenhouse environment. Aiming at narrowing the gap between the common context and the context of real life greenhouse conditions, we consider different data augmentation techniques and verify their effectiveness for the adaptation of the state-of-the-art few-shot learning models. Fine-grained descriptions are shown below.}

\suiyi{Concretely, taking an image $I$ (size of $w \times h$) as input, the generalized feature learner $D$ yields a set of meta features $F = D(I), F \in \mathbb{R}^{w\times h \times m}$. The feature re-weighter $M$ embeds images from the support images into a set of re-weighting vectors, \textsl{i.e.,} coefficients, $w_i$, and returns the category-relevant feature $F_i$ by }
\begin{equation}
    F_i= F \otimes w_i
\end{equation}
 \suiyi{where $i$ indicates a certain category, and $i = 1, ..., N$, and $\otimes$ is the channel-wise multiplication.  }

\suiyi{With the obtained $F_i$, it is then fed into a prediction model $P$, which returns the likelihood of object existence $o_i$, the predicted location of the object bounded by the offset of $(x, y, h, w)$, and a corresponding classification score $c_i$: }
\begin{equation}
  \{ o_i, x_i, y_i, h_i, w_i, c_i \} = P(F_i)
\end{equation}

\suiyi{A novel two-stage learning scheme was proposed~\cite{kang2019few} to guarantee the generalization of the model when dealing with new categories with few samples. }

\suiyi{\textbf{The first stage is the base learning stage.} During this stage, each base category is of abundant training samples with location labels. To ensure the model's capability of detecting target objects by making full use of a good re-weighting vector, $D, M, P$ are trained jointly. }

\suiyi{Formally, the base categories from the base training set were split into a set of few-shot detection task $T_j$. Each task was defined as $T_j = S_j \cup Q_j = \{ (I^j_1,M^j_1), ..., (I^j_N,M^j_N) \} \cup \{ (I^q_N,M^q_N) \} $, where $S$ is the support set containing $N$ samples from a different base category, and $Q_j$ is the Query set for evaluating the performance. Finally, the three modules are optimized jointly by minimizing the loss defined below: }

\begin{equation}
  \min_{\theta_D, \theta_M, \theta_P}  \sum_j %\mathcal{L}_{j}  = \sum_j \mathcal{L}_{det}(P_{\theta_p} (D_{\theta_D} (I^j_q)\otimes M_{\theta_M}(S_j)) ),
 \mathcal{L}_{j}  =  \sum_j \mathcal{L}_{det}(P_{\theta_p} (D_{\theta_D} (I^q_j)\otimes M_{\theta_M}(S_j)), M^q_j ),
\end{equation}
\suiyi{where $\theta_D $, $\theta_M $, and $\theta_P$ are the parameters of $D, M$ and $P$ respectively. $L_{det}$ is an adopted detection loss defined as the sum of (1) the cross-entropy loss over the calibrated category scores~\cite{kang2019few} (2) the bonding box regression loss~\cite{redmon2017yolo9000}, and (3) the objectiveness regression loss~\cite{redmon2017yolo9000}. }

\suiyi{\textbf{The second stage is the few-shot fine-tuning stage.} During this stage, the model is trained on both base and novel categories, including `cucumber', but only $k$ bounding boxes are available for each novel category. Similarly, only $k$ bounding boxes are included for the base categories. The training process is the same as the one in the first stage, but with less iterations and novel categories are involved. }

 %In order to get better performance, the dataset of base classes used for base training will be modified, aiming to reduce the gap between the common context and the context of greenhouse.

\suiyi{As pointed out in Section \ref{sec:intro} that there could be significant domain-shift between the categories of base categories and the target novel category in real life applications. As thus, we should transfer not only the detection knowledge from the base categories to the new ones, but also the context-knowledge (\textsl{e.g.,} noisy image conditions, occlusions \textsl{etc.}) within real life scenarios. }
 
\suiyi{ Therefore, in this study, we explore different simple data augmentation techniques to narrow the gap between the distributions of base categories and target novel category in real life cases, \textsl{i.e.,}  the `cucumber'. }
 
 \suiyi{The first stage of FS-FRW is crucial for the final performance, since the tuning stage relies heavily on the generalized features. Intuitively, data augmentation is one of the straightforward and simplest way to bridge the gap of the distributions between the base categories and the new categories that come from real life greenhouse scenarios. To this end,  we propose two data augmentation strategies that are dedicated for few-shot object recognition: }

\begin{itemize}
\item \suiyi{\textbf{Background replacement of base training images}: The backgrounds and image conditions of real life greenhouse images are relatively more complex compared to the ones in commonly used benchmark datasets like PASCAL VOC and COCO, as demonstrated in Fig~\ref{teaser}. Hence, we collected a set of greenhouse background images and randomly replace the backgrounds of the base training set with the collected ones. Examples of the new base training samples are depicted in Fig.~\ref{replace background}. As shown, only the foreground base objects were kept. By doing so, the model is forced to take into account the image conditions/ real life environment statistics. Some feature-dimension highlighted by the generalized feature extractor could therefore dedicated to the greenhouse contexts. In parallel, it is then possible for the feature re-weighter to focus on these relevant features and yield more robust re-weighted vectors.}

\begin{figure}[t]
\centering
\mbox{\parbox{1\textwidth}{
 \begin{minipage}[b]{0.24\textwidth}
 \subfloat[]
 {\label{ }\includegraphics[width=\textwidth]{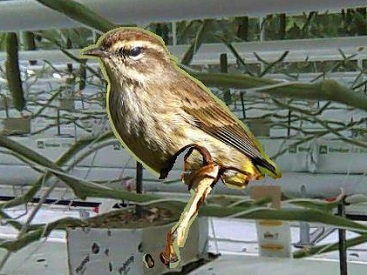}} \end{minipage}
 \begin{minipage}[b]{0.24\textwidth}
 \subfloat[]
 {\label{ }\includegraphics[width=\textwidth]{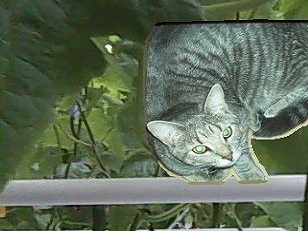}} \end{minipage}
}}
\caption{\suiyi{Examples of generated samples for base categories with replaced greenhouse background.}}
\label{replace background}
\end{figure}

\item \suiyi{\textbf{Adding `target background' as one of the base categories:} During the base training stage, an extra category `background' is added to the base dataset. This category could be adapted according to the specific task. In this study, it is the background images taken from the greenhouse. It is worth mentioning that the target objects do not show up in the images, and possible background regions of the target objects, \textsl{e.g.} regions of branches and leaves without cucumber in the greenhouse, were annotated as the ground truth. By including the task-relevant `background' as one of the base category, the model is then able to obtain important meta features that is related to the image conditions/environment of the target task and the feature re-weighter is able to associate these relevant-features to the target context.}

%By doing this, the base model already learns to highlight features specific to greenhouse and cucumbers, while the reweighting module learns to associate these features to greenhouse images. When tuning more precisely on cucumbers, the feature extractor will also have to highlight these specific features while the reweighting module will associate them with cucumbers. In a sens, most of the training task for cucumbers is done by training the model on "greenhouse background". The advantage of the "greenhouse background" class is that it doesn't need to be labeled, because the whole image is labeled as "background".
\end{itemize}

\suiyi{In a nutshell, both of these two augmentation approaches enrich the meta-training procedure so that it has access to the context (\textsl{e.g.} the image conditions, environment) without exposing the novel object itself. More implementation details of the two strategies are provided in Section~\ref{sec:exp}.} 
%explain why we can get higher precision by this two approach
%Through these two approaches, the features concerning the greenhouse context will be taken into consideration by the system during base training. Therefore, when it comes to tuning with only few examples of cucumber in greenhouse, the results are improved. 

\section{Experiments}
\label{sec:exp}
\suiyi{In this section, (1) the performances of the state-of-the-art few-shot object detection model FS-FRW in real life greenhouse applications are reported; (2) different data augmentation strategies were compared.  }

\subsection{Experimental Setup}
\suiyi{\subsubsection{Datasets }}
\suiyi{ Similar to the setup in~\cite{kang2019few}, The PASCAL VOC 07~\cite{everingham2010pascal}, and 12 train/val~\cite{everingham2010pascal} set was utilized for training while the VOC 07 test set was employed for testing. Following the same evaluation setting, among the 20 categories in the dataset, 15 of them were randomly selected as the base categories in the first stage for training generalized feature extractor, and the remained were considered as the novel ones. }

\suiyi{In the second stage for training the feature re-weighter, a very small set of training images were kept so that each category of objects only contain $k$ annotated bounding boxes for $k-$shot object detection problem. It has to be emphasized that we replaced one of the novel categories considered in~\cite{kang2019few} by our target `cucumber'. Details of base and novel category are summarized in Table~\ref{tab:cat_split}}.

\begin{table}[!htp]
\centering
\caption{Summary of base \& novel categories in the experiments.}
\begin{tabular}{|c|c|}  \hline
  \multirow{4}{*} {Base category} &  aeroplane, bicycle, boat, bottle,  \\  
& car, cat, chair, dining-table, \\  
    &  dog, horse, person, potted-plant,\\  
        &  sheep, train, tv-monitor\\ \hline 
 Novel category & bird, bus, cow, motobike, \textbf{cucumber} \\ \hline
\end{tabular}
\label{tab:cat_split}
\end{table}

%The goal of our two approaches is to reduce the gap between a common dataset used for the base training and our cucumber dataset.
%The two approaches are tested on PASCAL VOC common dataset.
%PASCAL VOC2012 and VOC2007 are gathered so that VOC2012 train, VOC2012 test, VOC2007 train and VOC2007 validation images are used to train the model. The validation steps are made on VOC2007 test images. This represents about 17 000 images for the training set and about 5000 images for the test set. These images belong to 20 classes. As in the Bingyi Kang  \& al article \cite{kang2019few}, 15 classes are used to train the base model and 5 of them are left to finetune the model : 

%The cucumber class is the only class that does not belong to PASCAL VOC dataset, since it is the class of interest, from our cucumber dataset. 

\subsubsection{Data Augmentation}
\suiyi{To bridge the gap between the context-knowledge of the scenarios used for training, and the ones of the target new scenarios in real life, in this study, we explored and examined not only the proposed data augmentation strategies presented in Section~\ref{sec:model}, but also the commonly used strategies including: }

\begin{itemize}
\item \suiyi{\textbf{Background Replacement of base training images  (BR):} Inspired by the data augmentation method proposed in~\cite{ling2019gans}, the pixel-wise segmentation masks provided by PASCAL VOC were utilized to extract the foreground semantic objects and the extracted objects were further posed on the greenhouse backgrounds. } 

\item \suiyi{\textbf{Adding `Target Background' as one of the base categories (ATB):} To implement this strategy, we simply replaced one of the base category `airplane' with the manually collected `greenhouse background' as described in Section~\ref{sec:model}. }

\item \textbf{Illuminance Adjustment (IA):} The illuminance of all the images of categorises except for the `cucumber' were simply enhanced using gamma correction by a factor of 1.5. 

\item \textbf{Contrast Adjustment (CA):} For this strategy, we increased the contrast of images belong to all categories except for the `cucumber' via contrast stretching with a factor of 2.

\end{itemize}

%\textbf{"Replace-background" approach settings}
%The first approach consists in replacing backgrounds of some images with backgrounds of cucumbers.
%PASCAL VOC comes with segmentation masks for some of its images. The segmentation masks were used to extract segmented objects from images and paste them on greenhouse backgrounds.
%3600 backgrounds of common images have been replaced among all the images in the common dataset. 

%\textbf{"Add-background-class" approach settings}
%The second approach consists in adding the class "greenhouse background" to the base training. To do so, the class "aeroplane" from PASCAL VOC has been replaced by the "greenhouse background" class.

%\textbf{Tuning details}
%Training on novel classes is a 10 shots tuning. It means that only 10 objects (not 10 images, potentially less) from each classes are used as training examples. 

\subsection{Experimental Results}
\suiyi{The Mean Average Precision (mAP) \modification{with IoU threshold of 50\%} was calculated to evaluate the performances of considered models\footnote{\suiyinew{Different batch size was set in this study, therefore, the reported results are lower compared to the ones in \cite{kang2019few} .}}. According to \cite{kang2019few}, FS-FRW achieves best performances among the compared models, it is thus considered as the baseline model. The 10-shots performance on the novel categories is reported in the TABLE~\ref{tab:re-base}.} It is obvious that the performance of FS-FRW on the target `cucumber' is worse than most of the other novel categories, better solutions are required to narrow the gaps.
Note that `bird' class achieves even lower performances than `cucumber', which was already the case as reported in~\cite{kang2019few} on PASCAL VOC. Most of bird images may be captured in wild and complex backgrounds. Such contexts also present a significant gap between most of the base classes, such as `aeroplane', `bike', \textsl{etc.}, which hinders the few shots tuning on this class. 

%\textcolor{red}{[??>>to add when server available]} images from our cucumber dataset were used as test images to get Mean Average Precision (mAP) of feature reweighting model on cucumbers once tuned. We compared mAP of the three experiments, respectively using original common dataset, using common dataset modified by "Replace-background" approach and using common dataset modified by "Add-background-class" approach. 
%Results of the there experiments are shown in the Tab.~\ref{result}

\begin{table}[!htbp]
\caption{10-shot performance (mAP \%) of the 5 novel categories.}
\begin{center}
\begin{tabular}{|c|c|c|c|c|} \hline 
  bird & bus & cow & motobike & \textbf{cucumber} \\ \hline 
15.66 & 41.06 & 27.24 & 35.59 & 27.00\\ \hline 
\end{tabular}
\label{tab:re-base}
\end{center}
\end{table}

\suiyi{As highlighted in Section~\ref{sec:intro}, one of the goals of this study is to explore different data augmentation methods, so that the state-of-the-art few-shot detection models could be better adjusted for real life cases. To this end, we compare the performance of FS-FRW equipped with different data augmentation approaches. Results are shown in TABLE~\ref{tab:DA_result}.}
\begin{table}[!htbp]
\caption{Performance comparison using different data augmentation strategies. }
\begin{center}
\begin{tabular}{|c|c|c|}
    \hline \textbf{Approach} & \textbf{mAP for `cucumber' (\%)}\\ 
    \hline  FS-FRW  & 27.00\\
    \hline  FS-FRW + BR & 29.33\\
    \hline FS-FRW + ATB  & \textbf{33.12}\\
     \hline FS-FRW + IA  &  14.04 \\
    \hline FS-FRW + CA  &  23.69 \\
     \hline
\end{tabular}
\label{tab:DA_result}
\end{center}
\end{table}

\suiyi{It could be observed that the two proposed strategies, \textsl{i.e.} BR and ATB, outperform the others. We also note that ATB performs slightly better than BR, one of the possible reasons is that by taking the task-relevant context into account earlier in an early stage (base learning training stage) is more efficient. Because the re-weighter fine-tuned in the second stage focus more on the characteristics of the object instead of the context of the real life applications. }

%Our two approaches improve the mean average precision of the feature reweighting model tuned on cucumbers, compared to the traditionnal base training on original dataset. Both approaches achieve similar results. Add-background-class approach achieves a barely higher result. This may be because during base training with this approach, the feature rewighting model already learns to associate greenhouse features with images of greenhouse and cucumbers, while with Replace-background Approach, the feature reweighting model must associate these features with other objects.

\suiyi{Examples are shown in Fig.~\ref{replace background} to better explain why better improvements could be achieved via ATB. Ideally, the model should detect only cucumbers in the target/first row so that the machine could harvest them within a reachable distance. Therefore, any other cucumbers that appear elsewhere were not labled in our dataset as mentioned in Section~\ref{datasetC}. By taking backgrounds as one of the base categories, the base model to some extend captures the fore/background information of the context, and thus able to better detect objects in the foreground other than the unrelated ones in the background. }

%Trained with Add-background-class approach, the feature reweighting model correctly detects only first row cucumbers, while when trained on the orignal dataset, it forgets one cucumber in the first row and wrongly detects one in the second row. 

\begin{figure}[!htpb]
\centering
\mbox{\parbox{1\textwidth}{
 \begin{minipage}[b]{0.24\textwidth}
 \subfloat[]
 {\label{ }\includegraphics[width=\textwidth]{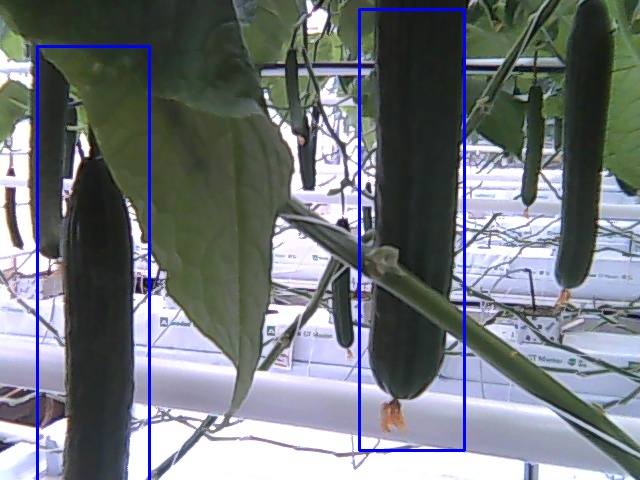}} \end{minipage}
 \begin{minipage}[b]{0.24\textwidth}
 \subfloat[]
 {\label{ }\includegraphics[width=\textwidth]{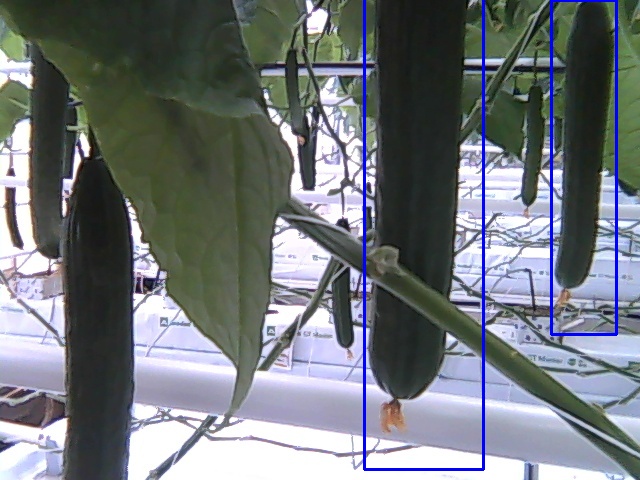}} \end{minipage}
}}
\caption{\suiyi{Examples explaining why performance could be improved by using ATB. (a) is the output of the feature re-weighter trained with ATB approach; (b) is the output of the feature re-weighter model trained without using special data augmentation method.}}
\label{visualResultImprov}
\end{figure}

\section{Conclusion \modification{and future work}}
\suiyi{In this study, in order to improve the robustness of few-shot object detection models for real life auto-harvest scenarios, 1) a `cucumber' dataset is released for the community; 2) the state-of-the art few-shot models is re-visited and employed for cucumber detection; 3) two novel data augmentation methods dedicated to real life few-shot applications are presented. Throughout experiments, it is verified that tested few-shot object detection model still need to be further improved to be applied in agricultural scenarios. The proposed data augmentation methods are proven to be effective.   }

\suiyinew{For} future work,\suiyinew{ 1) higher shot setting, \textsl{e.g.,} 15-shot results could be considered to further verify the impacts of the proposed strategies in low-shot regime; 2) more traditional data augmentation methods, \textsl{e.g.,} flipping, resizing, blurring \textsl{etc.}, could be compared and combined with the proposed ones to strengthen few-shot models in real life scenarios; 3) performance significant test, \textsl{e.g.} the one employed in \cite{ling2019prediction}, could be further utilized to ensure whether the performances are statistically improved.   } 

%\modification{\suiyinew{For} future work, we could compare the results of experiments with the number of shot increased, eg., 20-shot, 50-shot, 100-shot, etc., in order to see the number of shot needed to achieve an acceptable performance. Furthermore, we will compare our results with the traditional data augmentation strategies such as flipping, resizing, blurring, adding noise, etc., and expand our data augmentation methods with the combination of different data augmentation techniques. }

%\clearpage
%\section*{Acknowledgment}

%\section*{References}
\balance
\bibliographystyle{./IEEEtran}
\bibliography{ref}

\end{document}